\documentclass{article}
\usepackage{graphicx} 
\usepackage{blindtext}
\usepackage[accepted]{tmlr}

\usepackage{amsmath, amssymb, amsthm, mathtools}
\usepackage{siunitx}
\usepackage{booktabs, multirow, subcaption}
\usepackage{natbib}
\usepackage{xcolor}
\newcommand{\leo}[2]{\textcolor{gray}{#1}\textcolor{red}{#2}}

\usepackage{xcolor}

\newtheorem{proposition}{Proposition}
\usepackage{hyperref}

\title{Unbiased Open World Regularization for Fair Self-Supervised Learning}
\author{\name Léo Nicollier \email leo.nicollier@gmail.com \\
       \addr Université Paris-Saclay, CNRS, ENS Paris-Saclay, Centre Borelli, France \\
       Advanced Track and Trace
       \AND
       \name Marc Pic \\
       \addr Advanced Track and Trace
       \AND
       \name Pablo Musé \\
       \addr 
IIE, Facultad de Ingeniería, Universidad de la República, Uruguay\\ 
Université Paris-Saclay, CNRS, ENS Paris-Saclay, Centre Borelli, France
       \AND
       \name Enric Meinhardt-Llopis \\
       \addr Université Paris-Saclay, CNRS, ENS Paris-Saclay, Centre Borelli, France
       \AND
       \name Gabriele Facciolo \\
       \addr Université Paris-Saclay, CNRS, ENS Paris-Saclay, Centre Borelli, France \\  
       Institut Universitaire de France
       }\date{April 2026}

\begin{document}
\maketitle

\begin{abstract}
Despite recent advances, self-supervised learning (SSL) models and Joint-Embedding Predictive Architectures (JEPAs) remain susceptible to spurious correlations.
Many anti-collapse regularizers constrain the global representation distribution to match a prescribed target, such as a multivariate Gaussian or the uniform distribution on the sphere. 
However, global matching does not prevent groups defined by a bias attribute from occupying distinct regions of the representation space.
We show that existing encoder-only debiasing regularizers, including EnD and FSCL, can be interpreted as restricted forms of conditional distribution matching.
Building on this interpretation, we introduce Unbiased Open World (UOW), which matches the representation distribution within each observed bias group to a common target distribution.
At the population optimum, using a discriminative conditional discrepancy guarantees independence between the representation and the targeted attribute.
We instantiate UOW in Gaussian and spherical spaces and evaluate it across controlled synthetic settings and real-world fairness benchmarks.
Across Synthetic Engraving, Colored MNIST, CelebA, and Waterbirds, conditional matching substantially reduces bias compared with global regularization while preserving useful downstream representations.
Jointly conditioning on multiple bias attributes further reduces intersectional disparities missed by separate marginal constraints.
Finally, experiments on CelebA show that models can be trained from random initialization using UOW as a pretraining objective, provided checkpoints are selected according to fairness-aware validation criteria.
\end{abstract}

\begin{figure}[t]
    \centering
    \includegraphics[width=1\linewidth]{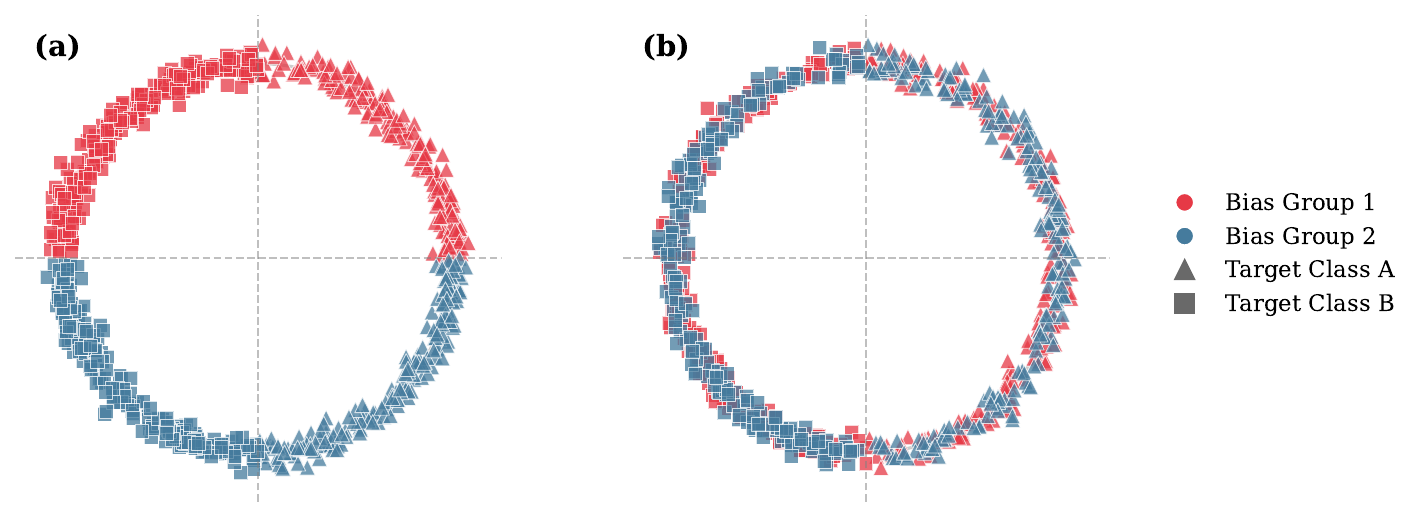}
    \caption{\textbf{Illustration of our distribution regularization framework instantiated on the sphere.}
    \textbf{(a) Global Regularization (Baseline):} Enforcing a global target distribution prevents representation collapse but allows spurious shortcuts to dictate the \textbf{latent structure}. The bias segregates the latent space, clustering distinct bias groups (colors) into isolated sub-regions.
    \textbf{(b) Unbiased Open World (UOW):} Shifting the constraint to \textit{conditional} distribution matching forces each bias group to independently match the target space (here, uniform on the sphere). Spurious features are blended, guaranteeing statistical independence ($X \perp\!\!\!\perp B$) while preserving the orthogonal separation of semantic targets (shapes).}
    \label{fig:teaser_geometry}
\end{figure}
\section{Introduction}
\label{sec:intro}

Self-supervised learning (SSL) has emerged as a dominant paradigm for extracting task-agnostic representations from complex visual data without relying on human annotations~\citep{infonce}. 
A fundamental challenge in SSL is preventing \textit{representation collapse}, where the encoder maps all inputs to a trivial constant~\citep{byol}. 
To prevent this, standard regularization techniques enforce a predefined global target distribution on the latent space, such as a multivariate Gaussian in Euclidean space~\citep{balestriero2025lejepaprovablescalableselfsupervised} or a uniform distribution on the unit sphere~\citep{spherejepa}.

Despite successfully preventing collapse, enforcing a global target distribution is insufficient when deep neural networks are applied to real-world data containing \textit{spurious correlations}. 
Instead of capturing the underlying causal semantic features, the models exploit the easiest predictive shortcuts available in the training distribution, such as background colors, dominant textures, or sensitive demographic attributes. 
Consequently, even a perfectly regularized latent space will implicitly segregate representations according to these dominant biases as in Figure~\ref{fig:teaser_geometry}, compromising both the fairness and the out-of-distribution generalization of the learned features. 
Addressing this entanglement via traditional approaches, such as adversarial min-max optimization~\citep{dann} or reconstructive generative models~\citep{VFAE}, introduces instability and hyperparameter sensitivity~\citep{locatello2019challenging, wgan}.
Alternatively, recent encoder-only methods~\citep{EnD, FSCL} successfully avoid generative modeling and minimax optimization by introducing dedicated debiasing regularizers.
While empirically effective, these approaches are viewed as distinct mechanisms. 


In this work, we first reformulate these empirical mechanisms under a unified principle, showing that they act as partial approximations of conditional distribution matching. 
Building upon this formalization, we propose Unbiased Open World (UOW), an encoder-only, distribution-agnostic framework designed to filter out known spurious attributes during pre-training.

We establish that, at the population optimum, matching every conditional representation distribution to a common target guarantees statistical independence between the representation and the targeted attributes (see Figure~\ref{fig:teaser_geometry}).

While our debiasing principle holds for any distribution, the specific target distribution dictates the utility of the representations for downstream tasks.
Recent theoretical work shows that the uniform distribution on the hypersphere ($\mathcal{U}(\mathbb{S}^{d-1})$) maximizes worst-case downstream performance for tasks such as $k$-nearest neighbors ($k$-NN) and linear ridge regression~\citep{spherejepa}. 
Therefore, to maximize downstream performance, we instantiate UOW primarily as \textit{conditional uniformity} on the unit sphere, using kernel-based statistical discrepancies that evaluate directly on the sphere, thereby bypassing the need for projections~\citep{spherejepa2}. 
We empirically validate this theoretical framework across spherical and Gaussian latent spaces, demonstrating successful bias mitigation for both target distributions, and confirming that the spherical constraint yields superior downstream performance.

We evaluate UOW across controlled synthetic settings and real-world fairness benchmarks.
We first introduce Synthetic Engraving, a new controlled dataset inspired by object biometrics~\citep{engraving}, in which a dominant DataMatrix structure provides a near-perfect shortcut.
In the most strongly entangled setting, conditional matching prevents subpopulation collapse and improves retrieval by $12.77$ percentage points.
On Colored MNIST, spherical UOW reduces the $k$-NN accuracy for predicting the spurious attribute from $100.0\%$ to $74.0\%$ while preserving target-task accuracy.
On CelebA and Waterbirds, UOW substantially reduces fairness violations relative to global regularization while maintaining competitive downstream accuracy.
We further show that jointly conditioning on multiple bias attributes mitigates intersectional disparities missed by separate marginal constraints.
Finally, experiments on CelebA demonstrate that UOW can be incorporated into pretraining from random initialization, provided that model selection is based on validation Equalized Odds.

In summary, our main contributions are as follows:
\begin{itemize}
    \item We provide a unified interpretation of prior encoder-only debiasing regularizers through conditional distribution matching: EnD promotes conditional second-order isotropy, while FSCL promotes conditional distributional spreading.
    \item We introduce UOW, a distribution-agnostic, encoder-only framework that matches each bias-conditional representation distribution to a common target. We show that, at the population optimum, a discriminative discrepancy guarantees independence between the learned representation and the targeted attribute.
    \item We introduce and release Synthetic Engraving, a new controlled dataset for object-biometric representation learning, in which a dominant DataMatrix structure provides a near-perfect shortcut that competes with subtle identity-specific micro-signatures.
    \item We evaluate UOW in Gaussian and spherical representation spaces across Synthetic Engraving, Colored MNIST, CelebA, and Waterbirds. UOW consistently reduces bias while preserving competitive downstream performance. We further show that jointly conditioning on multiple bias attributes reduces intersectional disparities missed when the attributes are constrained separately.
\end{itemize}

\section{Notation and Setup} \label{sec:notation}

Following the formalization of \cite{balestriero2025lejepaprovablescalableselfsupervised}, we consider a dataset composed of $N$ independent samples.
Each sample $x_n$ is associated with a discrete bias attribute $b_n \in \mathcal{B}$ (e.g., background color or a sensitive demographic trait), which we aim to disentangle from its semantic representation.
Each sample is observed through $V$ views, yielding data points $x_{n,v}\in\mathbb R^{D}$ for $n=1,\dots,N$ and $v=1,\dots,V$, where $D$ denotes the input dimension (e.g., for an image of spatial resolution $H \times W$ with $C$ channels, $D = C \times H \times W$).
These views  correspond to data augmentations of a given sample, such as geometric transformations or random crops.

Following standard SSL practice~\citep{SwAV, dinov1}, we distinguish between \textit{global} and \textit{local} views.
Global views are larger crops that preserve most of the image content, while local views are smaller crops capturing limited spatial context.
Among the total $V$ views, we denote the number of global views by $V_g$ and the number of local views by $V_l$ (such that $V = V_g + V_l$).
We index the views $v=1,\ldots, V$ such that the first $V_g$ indices correspond to the global views. 
We assume that the original samples $\{x_n\}_{n=1}^N$ are independent and identically distributed.

The official implementation of UOW is available at \url{https://github.com/darktanuki/UOW}.

\paragraph{Encoder and projection head.}
Let $f_\theta:\mathbb R^D\to\mathbb R^{d_h}$ denote the encoder.
For each view, its output is $h_{n,v}=f_\theta(x_{n,v})$.
A projection head $g_\phi:\mathbb R^{d_h}\to\mathbb R^d$ produces the representation
$z_{n,v}=g_\phi(h_{n,v})$ on which the training objectives act.
Downstream evaluation uses $h_{n,v}$ after discarding the projection head.

To match the chosen target distribution, the embeddings $z_{n,v}$ are mapped to their final representations $\tilde{z}_{n,v}$.
We define $\tilde{z}_{n,v} \coloneqq z_{n,v}$ for a Gaussian target in $\mathbb{R}^d$, whereas for a spherical target, we enforce an explicit $\ell_2$-normalization:
\begin{equation}
    \tilde z_{n,v} \coloneqq \frac{z_{n,v}}{\|z_{n,v}\|} \in \mathbb S^{d-1}.
\end{equation}

\section{Related Work}
\label{sec:related_works}

Our approach bridges recent advances in  self-supervised learning and fair representation learning.

\paragraph{Distribution Matching in Self-Supervised Learning.}
To prevent representation collapse, recent self-supervised frameworks enforce a predefined global target distribution on the latent space, such as a multivariate Gaussian~\citep{balestriero2025lejepaprovablescalableselfsupervised} or a uniform distribution on the unit hypersphere ($\mathcal{U}(\mathbb{S}^{d-1})$)~\citep{spherejepa}. 
This spherical uniform distribution is optimal for worst-case downstream performance in linear regression and $k$-nearest neighbors ($k$-NN)~\citep{spherejepa}. 
While initial methods enforced these targets via sliced estimators using random one-dimensional projections~\citep{balestriero2025lejepaprovablescalableselfsupervised, spherejepa}, they exhibit marginally higher variance and slightly slower convergence.
Consequently, subsequent works~\citep{zimmermann2025kerjepakerneldiscrepancieseuclidean, spherejepa2} introduced projection-free regularizers utilizing positive-definite kernels defined directly on the space.
We leverage these statistical measures, extending the global regularization framework to a conditional setting to explicitly mitigate known spurious biases.

\paragraph{Bias Mitigation and Fair Representations.}
Learning representations that are invariant to sensitive attributes is a long-standing objective in representation learning, dating back to Learning Fair Representations~\citep{zemel2013learning} (LFR).
A common formulation consists in reducing the Mutual Information (MI) between the learned representations and the sensitive attributes, for which various estimators have been proposed~\citep{cheng2020clubcontrastivelogratioupper}.
However, MI estimation remains challenging in high-dimensional settings~\citep{belghazi2021minemutualinformationneural}.
Several approaches rely on adversarial objectives~\citep{dann} or reconstructive generative models~\citep{VFAE}, which can introduce additional optimization complexity and hyperparameter sensitivity~\citep{wgan, locatello2019challenging}.
More recently, generative models, including diffusion-based approaches, have been used to improve dataset balance through data augmentation.
In contrast, our focus is on an encoder-only approach that operates directly in representation space, without requiring an additional data-generation stage.
Finally, post-hoc interventions (e.g., nullspace projection~\citep{NullItOut}) may be less effective when spurious correlations are already strongly embedded in the learned representations.


\paragraph{Supervised Debiasing Regularization.}
Several methods have explored the use of supervised regularization to mitigate spurious biases during representation learning. 
Among them, EnD~\citep{EnD} and FSCL~\citep{FSCL} achieve empirical bias mitigation without relying on generative modeling.
Conceptually, these approaches combine a task-specific objective based on target labels with an additional regularizer that leverages bias annotations to shape the representation space.
EnD penalizes inner products to encourage orthogonal subspaces among samples sharing the same spurious attribute.
Similarly, the repulsive component of FSCL's modified InfoNCE~\citep{infonce} objective balances contrastive repulsion across different bias groups.
While effective in practice, these methods primarily rely on local interactions between samples rather than explicitly enforcing a global distributional property of the latent space.
In Section~\ref{sec:method}, we formalize the relationship between these empirical methods, showing that both function as specific approximations of conditional distribution matching.

\section{Methodology}
\label{sec:method}

In this section, we formalize Unbiased Open World \leo{Regularization}{} (UOW\leo{Reg}{}), an encoder-only framework designed to disentangle spurious attributes during pre-training. 
We first define our self-supervised baseline, which balances multi-view invariance with a global target distribution.
We then demonstrate how shifting this geometric constraint to \textit{distribution matching} guarantees statistical independence from the bias. 
Lastly, we show that this probabilistic framework unifies prior debiasing regularizers.

\subsection{The Open World Baseline: Invariance and Global Uniformity}
To extract semantic representations without human annotations, self-supervised learning requires two opposing objectives: an alignment objective to capture task-agnostic features, and a regularization objective to prevent representation collapse~\citep{bardes2021vicreg}.

\paragraph{Invariant Prediction.} 
To capture semantic information, we enforce an alignment objective across the $V$ augmented views of a sample as in~\citep{balestriero2025lejepaprovablescalableselfsupervised, spherejepa}.
Given the embeddings $\tilde{z}_{n,v}$, we define the global prototype $\mu_n \coloneqq \frac{1}{V_g}\sum_{v=1}^{V_g}\tilde{z}_{n,v}$ as the average of its global views.
The invariance loss minimizes the squared Euclidean distance between all views and this prototype:
\begin{equation}
    \mathcal{L}_{\mathrm{inv}} = \frac{1}{V}\sum_{v=1}^{V} \mathbb{E}_n\|\mu_n - \tilde{z}_{n,v}\|_2^2.
\end{equation}

\paragraph{Global Distribution Matching.}
Minimizing $\mathcal{L}_{\mathrm{inv}}$ alone leads to representation collapse. 
To address this, we explicitly enforce the representations to match a predefined target random variable $Y$.
For a given view $v \in \{1, \dots, V\}$, let $X^{(v)}$ denote the empirical random variable obtained by sampling uniformly from the mini-batch representations $\{\tilde z_{n,v}\}_{n=1}^{N}$.
To prevent collapse, we minimize a statistical discrepancy measure $\mathcal{D}$ between the laws of $X^{(v)}$ and $Y$.
We refer to this objective as Open World Regularization (OWReg):
\begin{equation}
    \mathcal{L}_{\mathrm{OWReg}} = \frac{1}{V}\sum_{v=1}^{V} \mathcal{D}(X^{(v)}, Y).
\end{equation}

As shown by \citet{spherejepa2}, the choice of the discrepancy measure $\mathcal{D}$ dictates the structural properties of the representation space. 
Specifically, Maximum Mean Discrepancy (MMD) is well-suited for clustered spaces, whereas a Kernel Density Estimation (KDE)-based Kullback-Leibler (KL) divergence is more appropriate for continuous, unclustered data. 
This framework is agnostic to the probability law of the target random variable $Y$.
In the following empirical estimators, $X$ denotes the mini-batch random variable $X^{(v)}$, supported on $m=N$ representations.
We omit the view index when it is clear from context.
Following KerJEPA and SPHERE-JEPA, we multiply the empirical MMD by $m$ to normalize the scale of the test statistic across sample sizes.
We do not apply this scaling to the KL--KDE discrepancy, which is already expressed as an average log-density per sample.

To enforce a multivariate Gaussian target ($Y \sim \mathcal{N}(0, I)$ in $\mathbb{R}^d$), we compute the MMD using the kernel $k_{\mathcal{N}}$ introduced in KerJEPA~\citep{zimmermann2025kerjepakerneldiscrepancieseuclidean}. 
Because it avoids random projections, this formulation yields a lower-variance estimator that matches the objective of SIGReg~\citep{balestriero2025lejepaprovablescalableselfsupervised} in expectation. 
Following Theorems 7 and 8 from \citet{zimmermann2025kerjepakerneldiscrepancieseuclidean}, the empirical discrepancy for $X$ reduces to:
\begin{equation}
    \mathcal{D}_{\mathrm{MMD}}(X, \mathcal{N}(0, I)) = m\left(\mathbb{E}_{x,y\sim X}\left[k_{\mathcal{N}}(x, y)\right] - 2\, \mathbb{E}_{x\sim X}\left[ \int_{-1}^{1} h(\|x\|_2^2, u) \rho_d(u) du \right] + C_{\mathcal{N}} \right), \label{eq:mmd_gauss_explicit}
\end{equation}
where $h(r,u)=2^{-1/2}\exp(-ru^2/4)$ results from integrating the kernel against the Gaussian target, $\rho_d(u)=\frac{\Gamma(d/2)}{\sqrt{\pi}\Gamma((d-1)/2)} (1-u^2)^{(d-3)/2}$, for $u\in(-1,1)$ and $d\geq2$, is the density of the first coordinate of a uniformly distributed point on $\mathbb{S}^{d-1}$, and $C_{\mathcal N}=1/\sqrt{3}$ is the expected kernel value between two independent samples from the Gaussian target. 

Conversely, to enforce a uniform target on the hypersphere ($Y \sim\mathcal{U}(\mathbb{S}^{d-1})$), we evaluate the discrepancies on $X$ using the Heat Kernel $\varphi_t(x^\top y$) with scale parameter $t>0$, yielding the following deterministic closed-form estimators~\citep{spherejepa2}:
\begin{align}
    \mathcal{D}_{\mathrm{MMD}}(X, \mathcal{U}(\mathbb{S}^{d-1})) &= \frac{m}{C_{\mathrm{norm/MMD}}} \left( \mathbb E_{x,y\sim X}\left[\varphi_t(x^\top y)\right] - C_{\mathrm{bias/MMD}} \right), \label{eq:mmd_explicit} \\
    \mathcal{D}_{\mathrm{KL}}(X, \mathcal{U}(\mathbb{S}^{d-1})) &= \frac{1}{C_{\mathrm{norm/KL}}} \left( \mathbb E_{x\sim X} \left[ \log \mathbb E_{y\sim X_{-x}} \left[\varphi_t(x^\top y)\right] \right] - C_{\mathrm{bias/KL}} \right). \label{eq:kl_explicit}
\end{align}
Here, $X_{-x}$ denotes the leave-one-out empirical random variable obtained by sampling uniformly from the remaining $m-1$
representations after removing $x$.
The constants $C_{\mathrm{norm}}$ and $C_{\mathrm{bias}}$ normalize the underlying unscaled population discrepancy to be $1$ for a point mass and $0$ for the uniform distribution on $\mathbb S^{d-1}$.
The additional factor $m$ rescales only the empirical MMD test statistic and does not change its zeros.
Combining this distribution matching objective with the invariance loss defines our \textbf{Open World (OW)} baseline:
\begin{equation}
    \mathcal{L}_{\mathrm{OW}} = (1-\lambda) \mathcal{L}_{\mathrm{inv}} + \lambda\mathcal{L}_{\mathrm{OWReg}}.
\end{equation}

\subsection{Bias Mitigation via Conditional Distribution Matching (UOWReg)}

While the OW baseline ensures global distribution matching, it does not prevent representations from remaining statistically dependent on a spurious attribute $B$.
Indeed, distinct bias groups may occupy segregated regions of the target space while their mixture still matches the global target distribution (see Figure~\ref{fig:teaser_geometry}).
Consequently, global Open World regularization alone is insufficient to remove information about a known discrete bias variable $B\in\mathcal B$.

For each view $v$ and bias value $b\in\mathcal B$, let $X_b^{(v)}\coloneqq (X^{(v)}\mid B=b)$ denote the corresponding conditional random representation.
The following result concerns the population counterparts of the empirical random variables introduced above; we retain the same notation to avoid introducing additional symbols.
At the population level, matching every conditional distribution to a common target guarantees statistical independence:

\begin{proposition}[Independence via Conditional Matching]
\label{prop:independence}
Fix a view $v$.
If $X_b^{(v)}\sim Y$ for every $b\in\mathcal B$ such that $\mathbb P(B=b)>0$ then $X^{(v)}\sim Y$ and $X^{(v)}\perp\!\!\!\perp B$.
\end{proposition}

\textit{Derivation.}
Let $A$ be any measurable set in the representation space, and let $\mathbb P_Y$ denote the probability measure of the target distribution $Y$.
By the law of total probability, $\mathbb P(X^{(v)}\in A)=\sum_{b\in\mathcal B} \mathbb P(X^{(v)}\in A\mid B=b) \mathbb P(B=b)$.
Since $X_b^{(v)}\sim Y$ for every $b\in\mathcal B$, we have $\mathbb P(X^{(v)}\in A\mid B=b)= \mathbb P_Y(A)$.
Thus, $\mathbb P(X^{(v)}\in A)=\sum_{b\in\mathcal B} \mathbb P_Y(A) \mathbb P(B=b)=\mathbb P_Y(A)$, which establishes that $X^{(v)}\sim Y$.
Consequently, $\mathbb P(X^{(v)}\in A,B=b)=\mathbb P(X^{(v)}\in A\mid B=b) \mathbb P(B=b)=\mathbb P(X^{(v)}\in A) \mathbb P(B=b)$.
Therefore, $X^{(v)}\perp\!\!\!\perp B$.
\hfill $\square$

Following Proposition~\ref{prop:independence}, we introduce the \textbf{Unbiased Open World (UOW)} framework.
Its conditional distribution-matching regularizer, denoted UOWReg, is defined as:
\begin{equation}
    \mathcal L_{\mathrm{UOWReg}}
    =
    \frac{1}{V}
    \sum_{v=1}^{V}
    \frac{1}{|\mathcal B|}
    \sum_{b\in\mathcal B}
    \mathcal D
    \left(
        X_b^{(v)},
        Y
    \right).
    \label{eq:uowreg}
\end{equation}
For the empirical objective, $\mathcal B$ contains the bias groups represented by at least two samples in the mini-batch.
At the population level, it contains every bias value with positive probability.

The empirical estimators above apply with $m=N_b$, where $N_b$ is the number of mini-batch samples with bias value $b$.

The complete UOW training objective combines conditional distribution matching with view invariance and the global Open World anchor:
\begin{equation}
    \mathcal L_{\mathrm{UOW}}
    =
    (1-\lambda)\mathcal L_{\mathrm{inv}}
    +
    \lambda
    \left(
        (1-\alpha)\mathcal L_{\mathrm{UOWReg}}
        +
        \alpha\mathcal L_{\mathrm{OWReg}}
    \right).
    \label{eq:uow}
\end{equation}
The parameter $\lambda\in[0,1]$ balances view invariance and distribution matching, while $\alpha\in[0,1]$ interpolates between conditional and global matching.

Under exact conditional matching, the global term is theoretically redundant at the population level.
In practice, conditional discrepancies are estimated from smaller sub-batches and may yield noisy gradients.
Groups containing a single sample are excluded from UOWReg but remain included in the global Open World regularizer.
The global term therefore provides an additional geometric signal estimated from the complete mini-batch, which empirically stabilizes training.

We investigate the interaction between $\lambda$ and $\alpha$ in Section~\ref{sec:ablation}.
Unless otherwise stated, we use $\lambda=5\times10^{-2}$ and $\alpha=0.5$.
For the KL-regularized representations in Section~\ref{sec:experience}, we use $\lambda=1/3$ and $\alpha=0.25$.

For multiple annotated bias attributes, we treat their joint assignment as a single categorical bias variable $\mathbf B=(B_1,\ldots,B_M)$ and apply Equation~\ref{eq:uowreg} unchanged to its observed values $\mathbf b$.
For example, the binary attributes \texttt{Male} and \texttt{Young} define four intersectional groups $\mathbf b\in\{(0,0),(0,1),(1,0),(1,1)\}$, with $X_{\mathbf b}^{(v)}\coloneqq X^{(v)}\mid\mathbf B=\mathbf b$.
At the population level, matching every intersectional distribution $X_{\mathbf b}^{(v)}$ to $Y$ guarantees independence from the joint bias vector, and therefore from each individual attribute and their intersectional combinations.
In practice, this extension requires sufficiently populated intersectional groups to estimate their conditional discrepancies reliably.

\subsection{Theoretical Relationships with Prior Bias-Mitigation Methods}
\label{sec:unifying}

EnD~\citep{EnD} and FSCL~\citep{FSCL} combine target-supervised learning with bias-aware regularization.
As detailed in Appendix~\ref{app:prior_methods}, their objectives take the generic form
\begin{equation}
    \mathcal L
    =
    \mathcal L_{\mathrm{supervised}}
    +
    \gamma\mathcal L_{\mathrm{reg}},
\end{equation}
where $\gamma$ controls the regularization strength.
Their bias-aware regularizers can both be related to the conditional structure of UOWReg, but at different levels.
EnD acts as a restricted conditional second-order surrogate, whereas the repulsive component of $\mathrm{FSCL}^{\dagger}$ corresponds to a conditional KL--KDE instance of UOWReg.

\paragraph{EnD as a Conditional Second-Order Surrogate.}
The disentangling term $R_{\perp}$ of EnD penalizes absolute inner products between normalized representations sharing the same bias value.
Its population counterpart is
\begin{equation}
    \mathcal D_{\mathrm{EnD}}(X_b^{(v)}, \mathcal{U}(\mathbb{S}^{d-1})) \coloneqq \mathbb E_{x,\bar x\overset{\mathrm{i.i.d.}}{\sim}X_b^{(v)}} \left[ \left| x^\top\bar x \right| \right].
\end{equation}
Replacing $\mathcal D$ with $\mathcal D_{\mathrm{EnD}}$ in $\mathcal L_{\mathrm{UOWReg}}$ recovers $R_{\perp}$ under matched view and group weighting.
Replacing the target-supervised components of EnD with our invariance loss yields the comparison baseline denoted \textsc{EnD-SSL}, with $\alpha=0$.
As shown in Proposition~\ref{prop:end-second-order} (Appendix~\ref{app:end}), this surrogate promotes conditional second-order isotropy but does not guarantee independence from the bias attribute.

\paragraph{$\mathrm{FSCL}^{\dagger}$ as a Conditional KL--KDE Instance.}
$\mathrm{FSCL}^{\dagger}$, the self-supervised variant of FSCL introduced in~\citep{FSCL}, consists of an attractive term between two augmented views and a repulsive term between representations sharing the same bias value.
Its relationship with UOW is detailed in Appendix~\ref{app:fscl}.
For normalized representations, define
$k_\tau(x,\bar x)\coloneqq\exp(x^\top\bar x/\tau)$.
Let $X_{b,-x}^{(v)}$ denote the leave-one-out empirical random variable obtained from $X_b^{(v)}$ after removing the anchor $x$.
The empirical repulsive term can be written as
\begin{equation}
    \mathcal D_{\mathrm{FSCL}^{\dagger}}
    \left(
        X_b^{(v)},
        \mathcal U(\mathbb S^{d-1})
    \right)
    \coloneqq
    \mathbb E_{x\sim X_b^{(v)}}
    \left[
        \log
        \mathbb E_{y\sim X_{b,-x}^{(v)}}
        \left[
            k_\tau(x,y)
        \right]
    \right].
\end{equation}
The inner expectation is a leave-one-out kernel density estimate of the conditional representation distribution.
Up to kernel normalization and additive constants, the resulting average log-density estimates $\mathrm{KL}(X_b^{(v)}\|\mathcal U(\mathbb S^{d-1}))$.
Replacing the empirical conditional discrepancy in $\mathcal L_{\mathrm{UOWReg}}$ with this KL--KDE estimator recovers the repulsive component of $\mathrm{FSCL}^{\dagger}$.
Since its attractive component is proportional to our invariance loss, the complete $\mathrm{FSCL}^{\dagger}$ objective corresponds to a particular UOW objective with $\alpha=0$ and a fixed $\lambda$, as formalized in Proposition~\ref{prop:fscl-dagger-equivalence} (Appendix~\ref{app:fscl}).

\paragraph{Generalizing Conditional Bias Mitigation.}
These relationships provide different levels of conditional distributional control.
EnD promotes conditional second-order isotropy, whereas the KL--KDE regularizer of $\mathrm{FSCL}^{\dagger}$ captures richer distributional information.
UOW makes the conditional discrepancy explicit and allows it to be selected according to the structure of the representation space.
When the corresponding population discrepancy is discriminative and every conditional distribution is matched to the common target, Proposition~\ref{prop:independence} guarantees $X^{(v)}\perp\!\!\!\perp B$.

\section{The Synthetic Engraving Task}
\label{sec:synthetic_engraving}

To evaluate the ability of self-supervised representations to disentangle global structures from fine-grained signals, we introduce \textbf{Synthetic Engraving}, a synthetic dataset. While standard demographic bias benchmarks (e.g., CelebA) involve a limited number of attributes, this setting increases the complexity: a wide variety of structured macro-patterns acts as a spurious correlation that masks subtle, identity-defining features.

\subsection{Industrial Motivation}
In industrial applications such as physical authentication, manufactured components are often engraved with global tracking identifiers~\citep{engraving} (e.g., DataMatrix codes). While these codes provide traceability, physical authentication relies on extracting unique microscopic variations inherent to the material or the engraving process.

However, the high-contrast macro-structure of the tracking code often acts as a predictive shortcut for standard neural networks. This leads the latent space to collapse around the macro-structure, causing the network to ignore the task-relevant micro-signatures. This phenomenon is analogous to demographic biases in human biometrics, providing a controlled proxy task to evaluate subpopulation collapse.

\subsection{Dataset Construction}
\begin{figure}[t]
    \centering
    \includegraphics[width=\textwidth]{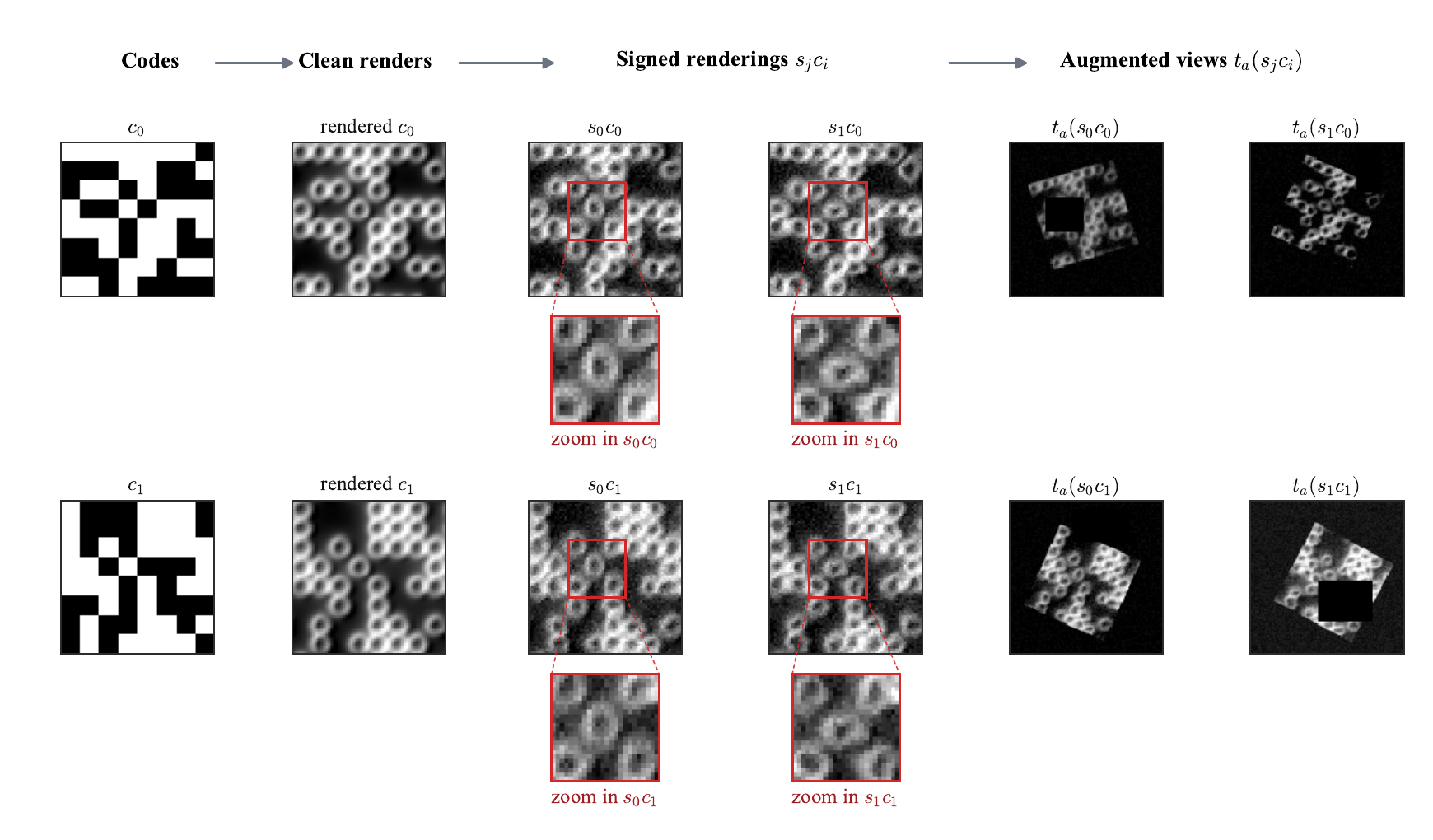}
    \caption{Overview of the Synthetic Engraving generation pipeline, which simulates a physical process where a binary data matrix is laser-etched onto a vibrating metal plate and photographed under lateral lighting.
    \textbf{Left to Right:} The abstract binary code $c_i$ is rendered into
    pixel space, injected with a unique structural imprint $s_j$ (the
    micro-texture) to form the signed rendering $s_j c_i$, and finally
    subjected to severe acquisition transformations $t_a$. \textbf{Rows \&
    Columns:} The top row is generated using code $c_0$ and the bottom row
    using $c_1$. Within the signed renderings block, columns share the same
    imprint ($s_0$ on the left, $s_1$ on the right).}
    \label{fig:generation_pipeline}
\end{figure}

To control the generative factors, our dataset decouples the global reference structure, the local structural identity, and the acquisition transformations (Figure~\ref{fig:generation_pipeline}).

Each sample is built upon a binary \emph{DataMatrix} $c_i \overset{\text{i.i.d.}}{\sim} \mathrm{Unif}(\{0,1\}^{8 \times 8})$. Unlike standard fairness benchmarks that feature a limited number of bias classes, our formulation introduces a large space of structured spurious correlations ($2^{64}$ possible macro-structures). This matrix dominates the image, creating a predictive shortcut analogous to a demographic attribute.

To each code $c_i$, we associate a random \emph{structural imprint} $s_j$.
The imprint is obtained by sampling a smooth random vector field and applying the induced deformation to the rendered image and adding a small amount of uniform noise. 
This process creates a unique microscopic texture characterizing the instance while preserving its macroscopic binary structure, yielding the composed object $s_j c_i$.

Finally, to simulate acquisition variability, we apply a set of transformations $t_a \in \mathcal{A}$ (rotation, translation, Gaussian blur, and random masking), resulting in the observed image $t_a(s_j c_i)$.

\subsection{The Bias Dominance Challenge}
\label{sec:challenge}

To evaluate representation learning under varying degrees of bias dominance, we control the number of distinct structural imprints $m$ associated with the \emph{same} global reference code during training. 

For a fixed dataset size $N$, the training set $\mathcal{D}_m$ is defined as:
\begin{equation}
    \mathcal{D}_m = \{\, t_a(s_{j_k} c_i) \mid i = 1,\dots,N/m,\; k=1,\dots,m,\; a \in \mathcal{A} \,\}.
\end{equation}
The parameter $m$ influences the task difficulty. A large $m$ provides sufficient intra-code variance for the network to identify $c_i$ as a static background and focus on $s_j$. Conversely, a small $m$ (e.g., $m=2$) creates a setting where the global structure $c_i$ is strongly correlated with the local variations.\footnote{The theoretical extreme $m=1$ yields a perfect one-to-one correlation between the global structure and local identity. Resolving this purely degenerate case without generative priors remains an open challenge and falls outside the scope of our encoder-only framework.}

\paragraph{Evaluation Metric.}
If a model is unconstrained, the code $c_i$ acts as a spurious shortcut, causing representations within this bias group to cluster together. Consequently, the network fails to distinguish the identities $s_j$. To measure this, we evaluate representation quality via cosine-similarity retrieval within the same macro-structure. Given a query image $x = t_a(s_j c_i)$, we rank all other samples sharing the same code $c_i$ and report the \textbf{Bias-conditional mAP ($\text{mAP}_{|c}$)}.

\section{Experiments}
\label{sec:experience}

We evaluate UOW from controlled synthetic settings to real-world fairness benchmarks.
We first use Synthetic Engraving to study fine-grained disentanglement under a controlled spurious correlation.
We then use Colored MNIST to visualize and quantify the effect of conditional distribution matching.
Finally, we evaluate fairness and downstream utility on CelebA and Waterbirds.
Additional experiments investigate intersectional conditioning, pretraining from random initialization, and the interaction between the regularization strength $\lambda$ and global-anchor weight $\alpha$.
Across all settings, we compare UOW with global Open World regularization and existing bias-mitigation methods.

\paragraph{Implementation Details.}
Unless otherwise stated, all methods within each experiment share the same backbone, projection head, augmentations, optimizer, batch size, and training schedule.
To maintain focus on our core analysis, we detail full architectural specifications, hyperparameter configurations, and extended dataset descriptions in Appendix~\ref{app:implementation}.

\subsection{Controlled Evaluation on Synthetic Engraving}

We evaluate UOW on the Synthetic Engraving dataset introduced in Section~\ref{sec:synthetic_engraving} using two pretrained encoders:
DINOv3 ConvNeXt~\citep{simeoni2025dinov3} and an ImageNet-supervised ResNet-50~\citep{he2016deep}.
Each image contains a dominant DataMatrix structure together with a subtle signature that must be preserved for retrieval.
We treat the DataMatrix code $c_i$ as the observed bias attribute.
Because the signature space is continuous and does not exhibit a clustered semantic structure, both OW and UOW target the uniform distribution on the sphere using the KL--KDE discrepancy with heat-kernel scale parameter $t=2/256$.

Table~\ref{tab:uow_ow_backbone_comparison} reports validation retrieval performance with a uniform target on the unit sphere
as the number of signatures per DataMatrix code varies.
At $m=2$, UOW improves $\mathrm{mAP}_{|c}$ from $78.15\%$ to $90.92\%$ with ConvNeXt, a gain of $12.77$ percentage points. 
It also improves performance with ResNet-50, from $48.02\%$ to $54.96\%$, a gain of $6.94$ percentage points.
Thus, conditioning on the DataMatrix code benefits retrieval with both encoders in this strongly entangled setting. 
Absolute performance is lower with ResNet-50; since the encoders also use different pretraining, these results do not isolate the effect of architecture.

The benefit of UOW is concentrated at small values of $m$.
For ResNet-50, the largest observed gap occurs at $m=4$, where UOW exceeds OW by $9.10$ percentage points. By $m=128$, the gaps have fallen to $0.14$ points for ConvNeXt and $0.24$ points for ResNet-50.
This pattern is consistent with conditional matching being most useful when the DataMatrix code provides a strong shortcut.

\begin{table}[t]
\centering
\small
\setlength{\tabcolsep}{5pt}
\caption{Validation retrieval $\mathrm{mAP}_{|c}$ (\%) on
Synthetic Engraving, with $m$ training signatures per
DataMatrix code. Both OW and UOW target the uniform
distribution on the unit sphere using KL--KDE
($t=2/256$, $\lambda=1/3$); UOW uses $\alpha=0.25$.
Values are means $\pm$ standard deviations over three seeds.
UOW is undefined at $m=1$.}
\label{tab:uow_ow_backbone_comparison}
\begin{tabular}{rcccc}
\toprule
& \multicolumn{2}{c}{DINOv3 ConvNeXt} & \multicolumn{2}{c}{ResNet-50} \\
\cmidrule(lr){2-3} \cmidrule(lr){4-5}
Signatures per code ($m$) & OW & UOW & OW & UOW \\
\midrule
1   & 15.76 $\pm$ 1.08 & --
    & 10.89 $\pm$ 0.12 & -- \\
2   & 78.15 $\pm$ 1.17 & 90.92 $\pm$ 1.20
    & 48.02 $\pm$ 2.41 & 54.96 $\pm$ 3.58 \\
4   & 90.34 $\pm$ 3.01 & 96.21 $\pm$ 0.52
    & 66.21 $\pm$ 1.13 & 75.31 $\pm$ 2.38 \\
8   & 93.75 $\pm$ 2.83 & 97.03 $\pm$ 0.73
    & 78.90 $\pm$ 1.23 & 83.67 $\pm$ 2.59 \\
16  & 97.30 $\pm$ 1.33 & 98.49 $\pm$ 0.25
    & 87.50 $\pm$ 1.34 & 89.73 $\pm$ 1.67 \\
32  & 97.29 $\pm$ 1.65 & 98.73 $\pm$ 0.39
    & 90.48 $\pm$ 0.62 & 92.02 $\pm$ 0.80 \\
128 & 98.70 $\pm$ 0.84 & 98.84 $\pm$ 0.70
    & 93.89 $\pm$ 0.81 & 94.13 $\pm$ 0.26 \\
\bottomrule
\end{tabular}
\end{table}

At $m=1$, each training signature is associated with a unique DataMatrix code, making the code a particularly easy shortcut for OW.
In the conditional retrieval evaluation, all 100 candidates share the same code; code information therefore cannot distinguish the matching signature.
Consistent with reliance on this shortcut, OW obtains only $14.76\pm1.08\%$ $\mathrm{mAP}_{|c}$ with ConvNeXt and $10.89\pm0.12\%$ with ResNet-50.
We do not evaluate UOW at $m=1$ because its conditional KL--KDE estimator requires more than one
training sample per code.

\subsection{Visualizing Bias Disentanglement on Colored MNIST}

\begin{figure}[t]
    \centering
    
    \begin{subfigure}{0.48\textwidth}
        \includegraphics[width=\textwidth]{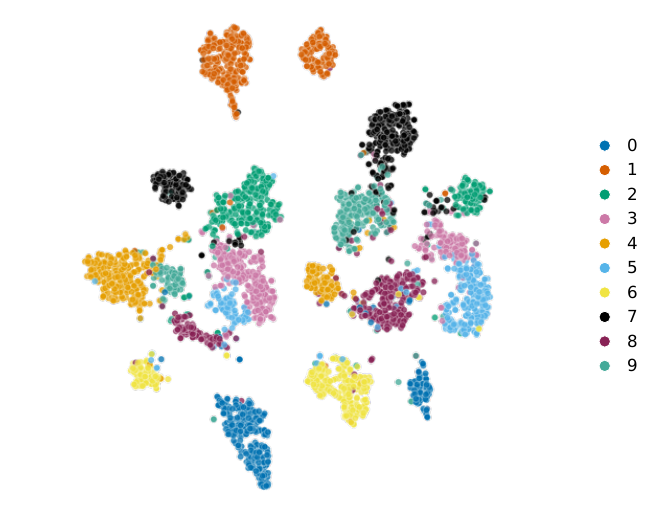}
        \caption{OW - Labels}
    \end{subfigure}
    \hfill
    \begin{subfigure}{0.48\textwidth}
        \includegraphics[width=\textwidth]{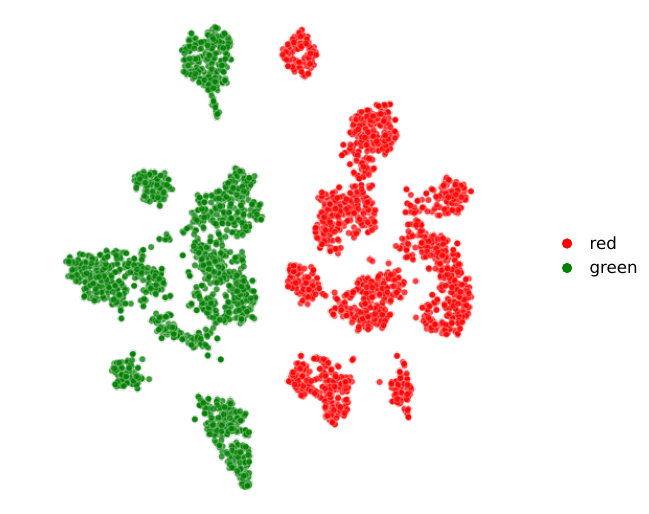}
        \caption{OW - Bias}
    \end{subfigure}
    
    \vspace{0.4cm} 
    
    \begin{subfigure}{0.48\textwidth}
        \includegraphics[width=\textwidth]{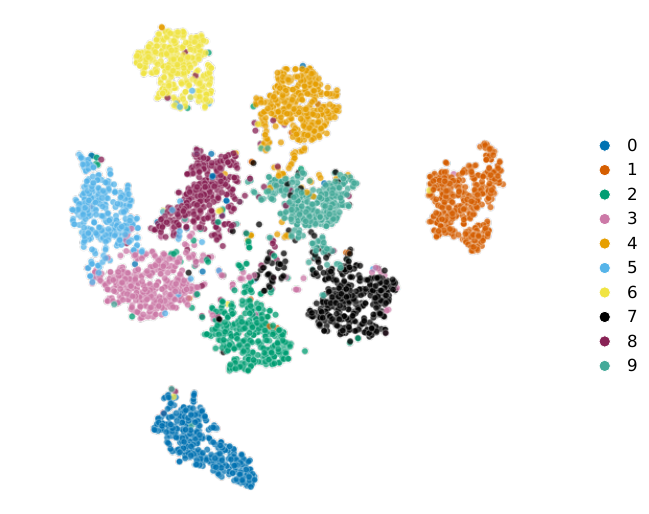}
        \caption{UOW - Labels}
    \end{subfigure}
    \hfill
    \begin{subfigure}{0.48\textwidth}
        \includegraphics[width=\textwidth]{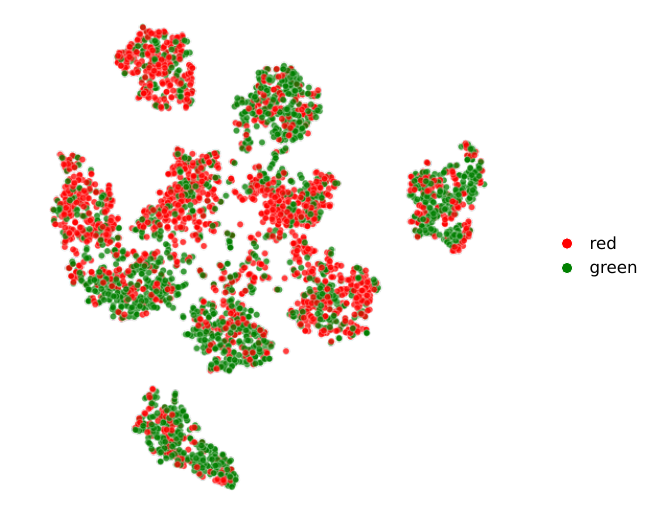}
        \caption{UOW - Bias}
    \end{subfigure}
    
    \caption{
    t-SNE visualization of Colored MNIST representations learned with a uniform target on the unit sphere, $\mathcal U(\mathbb S^{d-1})$, using MMD with the heat kernel. 
    Rows compare global OW (top) and conditional UOW (bottom); columns show digit identity (left) and the binary background color (right). 
    UOW preserves digit-class separation while increasing mixing between the two background colors. 
    Frozen $k$-NN bias accuracy decreases from $100.0\%$ to $74.0\%$ ($50\%$ for uniform random guessing), while digit accuracy remains comparable ($95.8\%$ for OW; $96.0\%$ for UOW). 
    }
    \label{fig:tsne_cmnist}
\end{figure}

We use Colored MNIST~\citep{cmnist} to visualize and quantify the effect of conditional distribution matching.
Digit identity is the semantic target, while background color, which takes two possible values, is the binary bias attribute.
Uniform random guessing of the background color yields $50\%$ accuracy.
Figure~\ref{fig:tsne_cmnist} shows two-dimensional t-SNE projections of the representations learned by OW and UOW, colored by digit identity (left column) or background color (right column).
Both methods target the uniform distribution on the unit sphere, $\mathcal U(\mathbb S^{d-1})$, using MMD with the heat kernel~\citep{spherejepa2}.
Table~\ref{tab:cmnist_knn} additionally reports results for a standard Gaussian target.
Under global OW regularization, the representations remain strongly organized by the dominant background color.
In contrast, UOW preserves the separation between semantic digit classes while producing substantially greater mixing between colors.
This qualitative result illustrates the geometric effect of conditional distribution matching, but does not by itself establish statistical independence.

To quantify the remaining information about the target and bias attributes, we evaluate the frozen representations using separate $k$-NN classifiers.
For both target distributions, the OW baseline yields $100.0\%$ bias accuracy, indicating that the background color remains perfectly predictable.
With a Gaussian target, UOW reduces bias accuracy to $70.2\%$ while increasing target accuracy from $94.7\%$ to $95.6\%$.
With a spherical target, UOW reduces bias accuracy to $74.0\%$ while preserving target utility, with accuracy increasing from $95.8\%$ to $96.0\%$.
This confirms that UOW effectively mitigates the targeted bias without compromising semantic utility.

\begin{table}[t]
    \centering
    \caption{Frozen $k$-NN evaluation on Colored MNIST, comparing global OW and conditional UOW with standard Gaussian $\mathcal N(0,I_d)$ and uniform on the sphere $\mathcal U(\mathbb S^{d-1})$ targets. 
    The spherical variants use MMD with the heat kernel. 
    Target accuracy measures digit classification (higher is better); bias accuracy measures prediction of the binary background color (lower toward the $50\%$ uniform-random-guessing reference is better). 
    Results are means $\pm$ standard deviations over three seeds.
}
    \label{tab:cmnist_knn}
    \begin{tabular}{llcc}
        \toprule
        \textbf{Method} & \textbf{Target Distribution} & \textbf{Target Acc. ($\uparrow$)} & \textbf{Bias Acc. ($\downarrow$)} \\
        \midrule
        OW Baseline & Gaussian $\mathcal{N}(0, I)$ & 94.7 $\pm$ 0.8\% & 100.0 $\pm$ 0.0\% \\
        UOWReg (Ours) & Gaussian $\mathcal{N}(0, I)$ & \textbf{95.6 $\pm$ 0.2\%} & \textbf{70.2 $\pm$ 1.3\%} \\
        \midrule
        OW Baseline & Spherical $\mathcal{U}(\mathbb{S}^{d-1})$ & 95.8 $\pm$ 0.6\% & 100.0 $\pm$ 0.0\% \\
        UOWReg (Ours) & Spherical $\mathcal{U}(\mathbb{S}^{d-1})$ & \textbf{96.0 $\pm$ 0.1\%} & \textbf{74.0 $\pm$ 2.0\%} \\
        \bottomrule
    \end{tabular}
\end{table}

\subsection{Fair Representation Learning on CelebA and Waterbirds}

We evaluate UOW on CelebA~\citep{celeba} and Waterbirds~\citep{waterbirds}, two benchmarks exhibiting spurious correlations.
On CelebA, we consider two binary downstream tasks: predicting \emph{Attractive} or \emph{Big Nose}, with \emph{Male} as the observed binary bias attribute.
On Waterbirds, the task is to distinguish waterbirds from landbirds, while the observed bias distinguishes water from land backgrounds.
After representation learning, we freeze the encoder and train a logistic-regression classifier using target labels.
We report target classification accuracy and Equalized Odds
(EO), which measures disparities in class-conditional
prediction rates between bias groups.
Higher accuracy indicates greater task utility, while lower
EO indicates smaller fairness disparities.

We compare UOW with FSCL$+$ and FSCL$^\dagger$~\citep{FSCL}, as well as \textsc{EnD-SSL}, our target-label-free adaptation of EnD~\citep{EnD}.
FSCL$+$ uses both target and bias annotations, whereas FSCL$^\dagger$, \textsc{EnD-SSL}, and UOW use only bias annotations during representation learning.
For a controlled comparison, \textsc{EnD-SSL} follows the UOW training protocol but replaces conditional distribution matching with the EnD orthogonality surrogate and omits the global OW anchor.

We evaluate Gaussian and spherical UOW variants using MMD, which is suited to the clustered semantic structures of these datasets.
The spherical variant uses the heat kernel~\citep{spherejepa2}, while the Gaussian variant uses the Euclidean kernel introduced in KerJEPA~\citep{zimmermann2025kerjepakerneldiscrepancieseuclidean}.
For both benchmarks, we use $\lambda=0.05$ and $\alpha=0.5$.

Table~\ref{tab:table4} shows that UOW consistently reduces Equalized Odds relative to global OW regularization.
On CelebA, spherical UOW reduces EO from $31.4\%$ to $6.5\%$ for \emph{Attractive}, and from $27.0\%$ to $3.6\%$ for \emph{Big Nose}, while retaining $75.7\%$ and $81.5\%$ accuracy, respectively.
Although \textsc{EnD-SSL} reaches lower EO, it incurs substantially larger accuracy losses, resulting in a less favorable fairness--utility trade-off.
The Gaussian variant retains slightly higher accuracy, whereas the spherical variant achieves lower EO.

On Waterbirds, spherical UOW reduces EO from $52.6\%$ to $8.1\%$ while increasing accuracy from $73.4\%$ to $78.3\%$.
The Gaussian variant obtains comparable results, and both UOW variants outperform FSCL$^\dagger$ and \textsc{EnD-SSL} in both fairness and downstream accuracy.

\begin{table}[t]
\centering
\caption{Target classification accuracy (Acc., $\uparrow$) and Equalized Odds disparity (EO, $\downarrow$) on CelebA and Waterbirds, reported in percent.
CelebA uses \emph{Attractive} or \emph{Big Nose} as the target and \emph{Male} as the bias; Waterbirds uses bird type as the target and background type as the bias.
Representation-learning methods are evaluated using logistic regression on frozen encoder features.
Gaussian and spherical OW/UOW variants use MMD with, respectively, the KerJEPA kernel and the heat kernel, targeting $\mathcal N(0,I_d)$ and the uniform distribution on the unit sphere.
CE denotes supervised cross-entropy training.
Results are means $\pm$ standard deviations across independent runs.}
\label{tab:table4}
\resizebox{\linewidth}{!}{
\begin{tabular}{lcccccc}
\toprule
\multirow{3}{*}{Method} & \multicolumn{4}{c}{CelebA} & \multicolumn{2}{c}{Waterbirds} \\
\cmidrule(lr){2-5}\cmidrule(lr){6-7}
& \multicolumn{2}{c}{$T$: Attractive / $S$: Male} & \multicolumn{2}{c}{$T$: Big Nose / $S$: Male} & \multicolumn{2}{c}{$T$: Bird type / $S$: Background} \\
\cmidrule(lr){2-3}\cmidrule(lr){4-5}\cmidrule(lr){6-7}
& EO ($\downarrow$) & Acc. ($\uparrow$) & EO ($\downarrow$) & Acc. ($\uparrow$) & EO ($\downarrow$) & Acc. ($\uparrow$) \\
\midrule
\multicolumn{7}{l}{\textit{Supervised (Requires labels during training)}} \\
\midrule
CE & 21.4 $\pm$ 1.9 & 82.4 $\pm$ 0.3 & 24.6 $\pm$ 6.2 & \textbf{84.4 $\pm$ 0.7} & 33.8 $\pm$ 3.7 & \textbf{83.8 $\pm$ 2.2} \\
\midrule
\multicolumn{7}{l}{\textit{Supervised Debiasing (Requires labels during training)}} \\
\midrule
FSCL$+$ & 16.6 $\pm$ 0.5 & \textbf{82.5 $\pm$ 0.1} & 12.4 $\pm$ 1.5 & 83.6 $\pm$ 0.1 & 34.3 $\pm$ 5.8 & 79.4 $\pm$ 3.3 \\
\midrule
\multicolumn{7}{l}{\textit{Fully Unsupervised}} \\
\midrule
LeJEPA (Gaussian)       & 50.4 $\pm$ 0.8 & 75.9 $\pm$ 0.5 & 24.7 $\pm$ 2.4 & 79.7 $\pm$ 0.2 & 71.8 $\pm$ 3.4 & 60.9 $\pm$ 2.1 \\
Baseline OW (Gaussian)  & 32.0 $\pm$ 0.7 & 80.2 $\pm$ 0.2 & 28.3 $\pm$ 1.0 & 82.2 $\pm$ 0.0 & 52.8 $\pm$ 3.0 & 72.5 $\pm$ 1.4 \\
Baseline OW (Spherical) & 31.4 $\pm$ 0.8 & 80.1 $\pm$ 0.0 & 27.0 $\pm$ 0.6 & 82.2 $\pm$ 0.1 & 52.6 $\pm$ 1.9 & 73.4 $\pm$ 1.0 \\
\midrule
\multicolumn{7}{l}{\textit{Self-Supervised Debiasing (Requires bias labels only)}} \\
\midrule
FSCL$^\dagger$ & 20.4 $\pm$ 0.5 & 80.4 $\pm$ 0.1 & 19.1 $\pm$ 0.4 & 82.3 $\pm$ 0.0 & 47.2 $\pm$ 0.5 & 74.2 $\pm$ 0.1 \\
\textsc{EnD-SSL} & \textbf{6.0 $\pm$ 2.5} & 63.8 $\pm$ 1.5 & \textbf{0.1 $\pm$ 0.2} & 78.8 $\pm$ 0.0 & 26.2 $\pm$ 3.3 & 71.1 $\pm$ 0.9 \\
UOW (Gaussian) & 12.0 $\pm$ 0.5 & 77.6 $\pm$ 0.1 & 4.0 $\pm$ 1.0 & 81.9 $\pm$ 0.2 & 9.2 $\pm$ 3.1 & 78.4 $\pm$ 0.5 \\
UOW (Spherical) & 6.5 $\pm$ 0.9 & 75.7 $\pm$ 0.2 & 3.6 $\pm$ 1.5 & 81.5 $\pm$ 0.2 & \textbf{8.1 $\pm$ 1.6} & 78.3 $\pm$ 0.1 \\
\bottomrule
\end{tabular}
}
\end{table}
 
\subsection{Debiasing Multiple and Intersectional Attributes}
\label{sec:celeba_intersectional}

We now consider whether UOW can mitigate bias when more than one attribute is observed.
On CelebA, we use \emph{Attractive} as the downstream target and \emph{Male} and \emph{Young} as observed biases.
A representation can be independent of each bias attribute individually while still retaining information about their combination.
To address this possibility, we define $\mathbf B=(B_{\mathrm{Male}},B_{\mathrm{Young}})$ and apply UOWReg to its four intersectional groups. 
If every conditional distribution $X_{\mathbf b}^{(v)}$ matches $Y$ at the population level, Proposition~\ref{prop:independence} implies $X^{(v)}\perp\!\!\!\perp\mathbf B$, and therefore independence from each attribute as well.

\begin{table}[t]
\centering
\caption{CelebA results with \emph{Attractive} as the downstream target and \emph{Male} and \emph{Young} as binary bias attributes.
OW and UOW target the uniform distribution on the unit sphere, $\mathcal U(\mathbb S^{d-1})$, using MMD with the heat kernel.
Separate conditioning gives equal weight to the Male-conditional and Young-conditional objectives; joint conditioning uses their four intersectional groups.
Acc. measures target classification accuracy; EO Male, EO Young, and EO Joint measure Equalized Odds disparities across Male groups, Young groups, and the four intersectional groups, respectively.
Values are percentages, reported as means $\pm$
standard deviations over three independent runs.}
\label{tab:celeba_intersectional}
\resizebox{\linewidth}{!}{%
\begin{tabular}{llcccc}
\toprule
Method & Bias conditioning & Acc. ($\uparrow$) & EO Male ($\downarrow$) & EO Young ($\downarrow$) & EO Joint ($\downarrow$) \\
\midrule
OW & Global only & \textbf{80.1 $\pm$ 0.1} & 31.2 $\pm$ 0.4 & 20.1 $\pm$ 0.5 & 29.0 $\pm$ 0.2 \\
\midrule
\multirow{3}{*}{UOW}
& Male only & 75.9 $\pm$ 0.5 & \textbf{6.4 $\pm$ 1.7} & 17.9 $\pm$ 2.0 & 13.0 $\pm$ 1.9 \\
& Male and Young separately & 77.7 $\pm$ 0.1 & 26.1 $\pm$ 1.5 & 15.7 $\pm$ 0.4 & 22.9 $\pm$ 1.0 \\
& Male $\times$ Young jointly & 75.2 $\pm$ 0.3 & 13.3 $\pm$ 2.3 & \textbf{11.5 $\pm$ 0.6} & \textbf{12.8 $\pm$ 0.5} \\
\bottomrule
\end{tabular}%
}
\end{table}

Table~\ref{tab:celeba_intersectional} reports results for spherical OW and UOW, targeting the uniform distribution on the unit sphere, $\mathcal U(\mathbb S^{d-1})$, using MMD with the heat kernel.
Conditioning UOW on \emph{Male} alone reduces Male EO from $31.2\%$ with global OW regularization to $6.4\%$.
Addressing \emph{Male} and \emph{Young} through separate conditional terms requires an additional choice of their relative weights, which we fix equally here.
Tuning these weights might improve the separate variant, but matching the two marginal distributions still does not guarantee independence from the joint attribute. 
Joint conditioning directly targets the four intersectional groups without introducing this additional weighting choice.
It reduces \emph{Male}, \emph{Young}, and joint EO relative to OW, yielding 13.3\%, 11.5\%, and 12.8\%, respectively, at the cost of lower target accuracy.

\subsection{UOW as a Pretraining Objective}
\label{sec:celeba_uow_pretraining}

\begin{figure}[t]
\centering
\includegraphics[width=\linewidth]{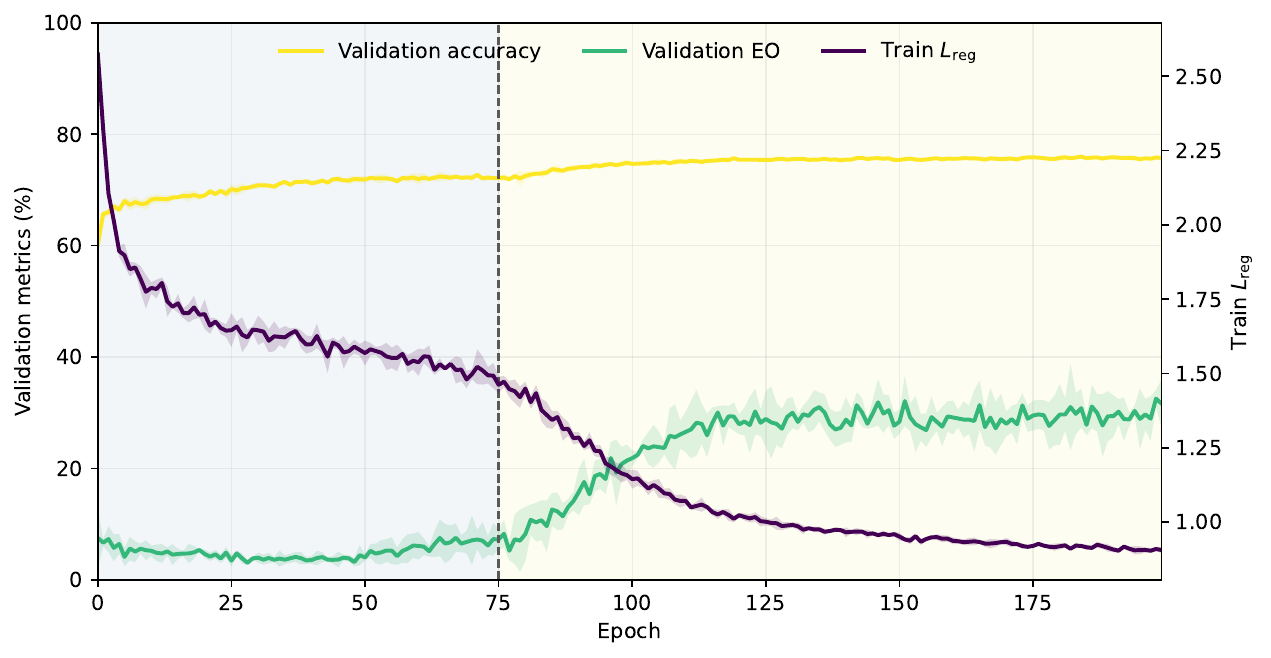}
\caption{From-scratch UOW pretraining dynamics on CelebA with $\alpha=0$.
Validation accuracy for \emph{Attractive} and EO with respect to \emph{Male} are measured by an online linear probe on the encoder representation (left axis); the training regularization loss is shown on the right axis.
Curves show means over four seeds. The dashed line marks epoch 75 as a reference checkpoint.}
\label{fig:celeba_pretraining_dynamics}
\end{figure}

UOW builds upon distribution-matching objectives developed for self-supervised pretraining.
We therefore test whether it can learn a useful, low-bias representation from random initialization on CelebA.

We train a randomly initialized ResNet-18 with $\lambda=0.05$ and $\alpha=0$.
The projection head $g$ is a two-layer MLP with Batch Normalization~\citep{balestriero2025lejepaprovablescalableselfsupervised,spherejepa}.
During training, an online linear probe measures \emph{Attractive} accuracy and EO with respect to \emph{Male} on the encoder output $h=f(x)$, without propagating gradients to the encoder.

Figure~\ref{fig:celeba_pretraining_dynamics} shows that accuracy and fairness evolve differently during pretraining.
At epoch 49, the encoder reaches 72.0\% accuracy and 3.2\% EO.
At the reference checkpoint marked at epoch 75, it reaches 72.1\% accuracy and 7.3\% EO.
By epoch 200, accuracy rises to 75.8\%, while EO reaches 31.7\%.
Thus, continuing to optimize the pretraining objective improves target accuracy but can substantially worsen fairness on the encoder output used downstream.

The early checkpoints show empirically that UOW can learn a useful, low-bias encoder representation from scratch.
They also show why checkpoint selection must account for fairness: selecting solely for accuracy would favor a much less fair representation.
One possible explanation for the later rise in EO is that the training objective acts on $z=g(f(x))$, while the probe evaluates $f(x)$ after discarding $g$.

\subsection{Interaction Between Regularization Strength and the Global Anchor}
\label{sec:ablation}

We study the joint effect of the total distribution-matching strength $\lambda$ and the global-anchor weight $\alpha$.
In Equation~\ref{eq:uow}, these parameters induce effective weights $\lambda(1-\alpha)$ for conditional matching and $\lambda\alpha$ for global matching.
Consequently, changing $\alpha$ not only modifies the relative contribution of the global anchor, but also changes the effective strength of the conditional constraint.
We examine this interaction in two complementary regimes: continuous, unclustered representations regularized with KL--KDE, and clustered representations regularized with MMD.
All configurations in this ablation use a uniform target distribution on the hypersphere and are evaluated using a single random seed.

\paragraph{Unclustered representations ($\mathcal D=\mathcal D_{\mathrm{KL}}$).}

We first consider the most strongly entangled Synthetic Engraving setting, with only $m=2$ signatures per DataMatrix code.
Because each conditional KL--KDE estimate is computed from a very small subpopulation, this setting exposes the stabilizing role of the global anchor.
Table~\ref{tab:engraving_lambda_alpha_ablation} reports the best validation $\mathrm{mAP}_{|c}$ obtained for each $(\lambda,\alpha)$ configuration.

\begin{table}[t]
\centering
\caption{
Synthetic Engraving ablation over the regularization strength $\lambda$ and global-anchor weight $\alpha$ in the $m=2$ regime.
Each cell reports the best validation $\mathrm{mAP}_{|c}$ (\%) obtained using a single seed.
}
\label{tab:engraving_lambda_alpha_ablation}
\begin{tabular}{rccccc}
\toprule
$\lambda\backslash\alpha$
& $0$
& $0.25$
& $0.5$
& $0.75$
& $1$ \\
\midrule
$1/4$ & 5.20 & 5.22 & 5.34 & 73.58 & 68.47 \\
$1/3$ & 7.50 & \textbf{90.69} & 80.45 & 77.80 & 77.03 \\
$1/2$ & 5.34 & 6.32 & 81.62 & 77.06 & 76.21 \\
\bottomrule
\end{tabular}
\end{table}

Purely conditional matching ($\alpha=0$) performs poorly for all three values of $\lambda$, consistent with the difficulty of estimating a KL--KDE discrepancy from very small conditional subsets.
Introducing a global component can produce a sharp improvement, although the amount required depends strongly on $\lambda$.
The best configuration, $(\lambda,\alpha)=(1/3,0.25)$, reaches $90.69\%$ $\mathrm{mAP}_{|c}$.
For $\lambda=1/4$ and $\lambda=1/2$, the best results instead occur at $\alpha=0.75$ and $\alpha=0.5$, respectively.
The purely global objective ($\alpha=1$) avoids the severe performance degradation observed under purely conditional matching, but remains below the best intermediate mixtures.
These results suggest that the global anchor mitigates the finite-batch estimation difficulties of conditional KL--KDE, while excessive global weighting weakens the conditional constraint.

\paragraph{Clustered representations ($\mathcal D=\mathcal D_{\mathrm{MMD}}$).}

\begin{figure}[t]
\centering
\includegraphics[width=0.75\linewidth]{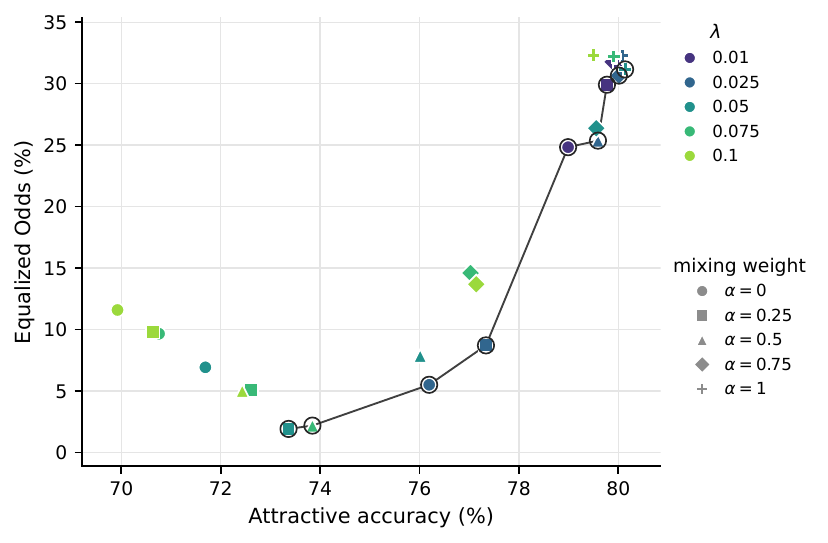}
\caption{
Seed-0 fairness--utility trade-off on CelebA across 25 configurations of the regularization strength $\lambda$ and global-anchor weight $\alpha$.
Colors identify $\lambda$, while markers identify $\alpha$.
Circled points connected by the black line are Pareto optimal: no other configuration simultaneously achieves higher \emph{Attractive} accuracy and lower Equalized Odds.
Configurations closer to the bottom-right provide a more favorable fairness--utility trade-off.
}
\label{fig:celeba_pareto}
\end{figure}

We next evaluate spherical MMD on CelebA, using \emph{Attractive} as the downstream target and \emph{Male} as the observed bias.
We consider $\lambda\in\{10^{-2},2.5\times10^{-2},5\times10^{-2}, 7.5\times10^{-2},10^{-1}\}$ and $\alpha\in\{0,0.25,0.5,0.75,1\}$.
Figure~\ref{fig:celeba_pareto} presents the resulting accuracy--Equalized Odds trade-off.

As shown in Figure~\ref{fig:celeba_pareto}, the fairness--utility trade-off depends jointly on $\lambda$ and $\alpha$.
Weak regularization, such as $\lambda=0.01$, preserves approximately $79$--$80\%$ accuracy but leaves EO above $24\%$, even under purely conditional matching.
At the other extreme, configurations with $\alpha=1$ recover the global OW objective and cluster around $79.5$--$80.1\%$ accuracy and $31.2$--$32.3\%$ EO.

Intermediate global-anchor weights provide access to substantially fairer operating points.
The configuration $(\lambda,\alpha)=(0.05,0.25)$ achieves the lowest EO, $1.91\%$, with $73.36\%$ accuracy.
Increasing utility along the frontier yields $73.85\%$ accuracy and $2.18\%$ EO at $(0.075,0.5)$, and $76.19\%$ accuracy and $5.50\%$ EO at $(0.025,0)$.
Notably, for $\lambda=0.05$, moving from $\alpha=0$ to $\alpha=0.25$ improves both accuracy and EO, suggesting that a moderate global anchor can improve the learned representation rather than merely trading fairness for utility.

Together, these seed-0 ablations suggest that the global anchor can mitigate finite-sample estimation difficulties, but that its optimal
contribution is not universal.
Conditional KL--KDE estimation on very small groups benefits strongly from a nonzero global component, whereas spherical MMD on CelebA admits several Pareto-optimal configurations with low or intermediate values of $\alpha$.
The regularization strength and global-anchor weight should therefore be selected jointly based on the representation structure, discrepancy estimator, and the desired fairness--utility operating point.

\section{Conclusion and Future Work}
\label{sec:conclusion}

In this work, we introduced Unbiased Open World (UOW), a distribution-agnostic framework for fair self-supervised learning.
By shifting the regularization objective from global to conditional distribution matching, UOW enforces statistical independence between learned representations and spurious attributes.
Our approach achieves this disentanglement through a stable, encoder-only architecture, bypassing the optimization instabilities of adversarial methods and the computational overhead of generative data augmentation.

Across Synthetic Engraving, Colored MNIST, CelebA, and Waterbirds, UOWReg substantially reduces the influence of observed biases relative to global Open World regularization while preserving useful downstream representations.
On Synthetic Engraving, it improves fine-grained retrieval when local signatures are strongly entangled with the DataMatrix code.
On CelebA and Waterbirds, it provides favorable fairness--utility trade-offs, while the intersectional experiment shows that joint conditioning can capture disparities missed by separate marginal constraints.
Our from-scratch experiment further demonstrates that UOW can serve as a pretraining objective, provided that training is stopped early based on validation EO.

\paragraph{Limitations, Broader Impact, and Future Work.}
Our formulation assumes that the targeted bias is known and that categorical bias annotations are available during training.
Joint conditioning also increases the number of groups exponentially with the number of attributes, making conditional estimation difficult for rare intersections.
Extending UOWReg to partially annotated, continuous, or dynamically discovered biases therefore constitutes an important direction for future work.
Moreover, fairness-aware checkpoint selection in our from-scratch experiment uses validation EO and consequently requires both target and bias labels for model selection, although these target labels are not used to optimize the encoder.
Finally, perfect disentanglement remains challenging when semantic and spurious variables are strongly correlated.
The perfectly correlated case ($m=1$) is unidentifiable without additional assumptions or priors and remains an open problem.

\subsection*{Acknowledgements}
This work was partially funded by ANRT with the CIFRE grant n° 2024/1221, Advanced Track and Trace, and Centre Borelli. 
This work was granted access to the HPC resources of IDRIS (Jean Zay supercomputer) under the allocation AD011017323 made by GENCI
The author also thanks Martin Jessenne for his contribution to the early development of this project during his internship.

\bibliography{references}

\appendix

\newpage

\section{Formulations of Prior Bias Mitigation Methods}
\label{app:prior_methods}

In Section~\ref{sec:unifying}, we write the objectives of prior methods as $\mathcal L=\mathcal L_{\mathrm{inv}}+\lambda\mathcal L_{\mathrm{reg}}$.
This section formalizes the relation between their regularization terms and the conditional objective of UOWReg.
We use the notation introduced in Section~\ref{sec:notation}.
In particular, $X_b^{(v)}$ denotes the representation associated with view $v$, conditionally on $B=b$.
A barred variable, such as $\overline X_b^{(v)}$, denotes an independent copy of $X_b^{(v)}$.

\subsection{Entangling and Disentangling (EnD)}
\label{app:end}

We relate the disentangling component $R_{\perp}$ of EnD~\citep{EnD} to the conditional regularization principle underlying UOWReg.
We focus on $R_{\perp}$, since the complete EnD objective also contains a supervised classification loss and an entangling component $R_{\parallel}$.

We first consider a single view and omit the view index for clarity.
For each $b\in\mathcal B$, let $x,\bar{x}\overset{\mathrm{i.i.d.}}{\sim}X_b\coloneqq X\mid B=b$.
Since EnD operates on $\ell_2$-normalized representations, $x,\bar{x}\in\mathbb S^{d-1}$.

We define the EnD surrogate for bias group $b$ as
\begin{equation}
    \mathcal D_{\mathrm{EnD}}(X_b,\mathcal U(\mathbb{S}^{d-1}))
    \coloneqq
    \mathbb E_{x,\bar{x}\overset{\mathrm{i.i.d.}}{\sim}X_b}
    \left[
        \left|x^\top\bar{x}\right|
    \right].
    \label{eq:end-surrogate}
\end{equation}
This surrogate measures the expected absolute correlation between two independent representations sharing the same bias value. Assuming that the EnD regularizer is applied separately to every view, its population counterpart can be written as
\begin{equation}
    \mathcal L_{\mathrm{reg}}^{\mathrm{EnD}}
    =
    \frac{1}{V}
    \sum_{v=1}^{V}
    \mathbb E_b
    \left[
        \mathcal D_{\mathrm{EnD}}(X_b^{(v)}, \mathcal{U}(\mathbb{S}^{d-1}))
    \right].
    \label{eq:end-conditional-form}
\end{equation}
This objective has the same conditional aggregation structure as the conditional component of UOWReg. 
However, EnD uses the surrogate $\mathcal D_{\mathrm{EnD}}$ in place of the statistical discrepancy $\mathcal D(X_b^{(v)},Y)$.

\begin{proposition}[Conditional second-order isotropy promoted by EnD]
\label{prop:end-second-order}
Assume that every bias group receives strictly positive weight in $\mathbb E_b$. 
For every $b\in\mathcal B$ and $v\in\{1,\ldots,V\}$, let $x,\bar{x} \overset{\mathrm{i.i.d.}}{\sim} X_b^{(v)}$ and define the conditional second-moment matrix $\Sigma_b^{(v)} \coloneqq \mathbb E_{x\sim X_b^{(v)}}[xx^\top]$.
Then
\begin{equation}
    \mathcal D_{\mathrm{EnD}}
    \left(
        X_b^{(v)},
        \mathcal U(\mathbb S^{d-1})
    \right)
    \geq
    \frac{1}{d}
    +
    \left\|
        \Sigma_b^{(v)}-\frac{1}{d}I_d
    \right\|_F^2.
    \label{eq:end-second-order-control}
\end{equation}
Consequently, the minimum possible value of $\mathcal L_{\mathrm{reg}}^{\mathrm{EnD}}$ is $1/d$.
If this minimum is attained, then $\Sigma_b^{(v)}=\frac{1}{d}I_d$ for every $b\in\mathcal B$ and every $v\in\{1,\ldots,V\}$.
Thus, when the EnD disentangling regularizer attains its population minimum, the conditional second moment is identical across bias values.
\end{proposition}

\begin{proof}
For $x,\bar{x}\overset{\mathrm{i.i.d.}}{\sim}X_b^{(v)}$, independence gives $\mathbb E \left[ (x^\top\bar{x})^2 \right] = \operatorname{tr} \left( \bigl(\Sigma_b^{(v)}\bigr)^2 \right)$.
Since $x\in\mathbb S^{d-1}$ almost surely, $\operatorname{tr}(\Sigma_b^{(v)}) = \mathbb E[\|x\|_2^2] = 1$.
It follows that $\operatorname{tr}\left( \bigl(\Sigma_b^{(v)}\bigr)^2 \right) = \frac{1}{d} + \left\| \Sigma_b^{(v)}-\frac{1}{d}I_d \right\|_F^2$.
Moreover, because $|x^\top\bar{x}|\in[0,1]$, $(x^\top\bar{x})^2 \leq |x^\top\bar{x}|$.
Therefore,
\begin{align}
    \mathcal D_{\mathrm{EnD}}
    \left(
        X_b^{(v)},
        \mathcal U(\mathbb S^{d-1})
    \right)
    &=
    \mathbb E
    \left[
        |x^\top\bar{x}|
    \right]
    \\
    &\geq
    \mathbb E
    \left[
        (x^\top\bar{x})^2
    \right]
    \\
    &=
    \frac{1}{d}
    +
    \left\|
        \Sigma_b^{(v)}-\frac{1}{d}I_d
    \right\|_F^2,
\end{align}
which proves~\eqref{eq:end-second-order-control}.
The lower bound $1/d$ is attainable when each $X_b^{(v)}$ is uniformly distributed over an orthonormal basis of $\mathbb R^d$.
Indeed, in this case $|x^\top\bar{x}|=1$ when the two draws select the same basis vector and is zero otherwise, giving an expectation of $1/d$.
Finally, if the average in~\eqref{eq:end-conditional-form} equals $1/d$, every conditional term with positive weight must attain its lower bound.
Equation~\eqref{eq:end-second-order-control} then implies $\Sigma_b^{(v)}=I_d/d$ for every $b$ and $v$.
\end{proof}

For the spherical target $Y\sim\mathcal U(\mathbb S^{d-1})$, a discriminative UOWReg discrepancy satisfies $\mathcal D(X_b^{(v)},Y)=0 \quad\Longrightarrow\quad X_b^{(v)}\sim Y.$
Consequently, at the population optimum, $\Sigma_b^{(v)}=I_d/d$ for every $b$ and $v$.
When minimized, the EnD surrogate recovers this second-order property, but it is not a discriminative discrepancy. 
For example, a distribution that is uniform over an orthonormal basis also satisfies $\Sigma_b^{(v)}=I_d/d$ and attains the minimum EnD value, despite not being uniform on the sphere. Different bias groups may therefore attain the same minimum using different conditional distributions.
Hence, EnD does not enforce $X_b^{(v)}\sim Y$ and does not guarantee $X^{(v)}\perp\!\!\!\perp B$.

The surrogate in~\eqref{eq:end-surrogate} is the population counterpart of the off-diagonal component of the EnD Gram regularizer.
The diagonal entries of the finite-sample Gram matrix are equal to one and therefore contribute only a representation-independent term.
Accordingly, up to this constant contribution and finite-sample normalization, EnD has the same conditional aggregation structure as UOWReg.
Its essential limitation is that it replaces the discriminative discrepancy of UOWReg with a surrogate that controls only conditional second-order isotropy.

\subsection{Fair Supervised Contrastive Learning (FSCL)}
\label{app:fscl}

Fair Supervised Contrastive Learning (FSCL)~\citep{FSCL} modifies the standard supervised contrastive objective by restricting the negative samples according to their sensitive attribute.
The original paper also introduces a self-supervised variant that does not require target labels, denoted $\mathrm{FSCL}^{\dagger}$ by its authors.

In $\mathrm{FSCL}^{\dagger}$, the attractive term aligns representations $x^{(v_1)}$ and $x^{(v_2)}$ obtained from two views of the same sample, while the denominator contains representations sharing the same bias value as the anchor.
For normalized representations, the similarity kernel is $k_\tau(x,y)\coloneqq\exp(x^\top y/\tau)$, where $\tau>0$ is the contrastive scale parameter.

Using the notation of Section~\ref{sec:method}, the repulsive component of $\mathrm{FSCL}^{\dagger}$ can be written, up to an additive constant independent of the representations, as
\begin{equation}
    \mathcal L_{\mathrm{reg}}^{\mathrm{FSCL}^{\dagger}}
    =
    \frac{1}{V}
    \sum_{v=1}^{V}
    \mathbb E_b
    \left[
        \mathbb E_{x\sim X_b^{(v)}}
        \left[
            \log
            \mathbb E_{y\sim X_{b,-x}^{(v)}}
            \left[
                k_\tau(x,y)
            \right]
        \right]
    \right].
    \label{eq:fscl-dagger-regularizer}
\end{equation}

The inner expectation is a leave-one-out kernel density estimate of the conditional representation distribution $X_b^{(v)}$.
Since $k_\tau(x,y)$ is proportional to $\exp(-\|x-y\|_2^2/(2\tau))$ for normalized representations, it corresponds to a Gaussian KDE on the sphere.
Up to kernel normalization and additive constants, the resulting average log-density estimates $\mathrm{KL}(X_b^{(v)}\|\mathcal U(\mathbb S^{d-1}))$.
Therefore, the repulsive component of $\mathrm{FSCL}^{\dagger}$ is a particular conditional KL--KDE instance of UOWReg.

The attractive component of $\mathrm{FSCL}^{\dagger}$ is proportional to the two-view invariance loss used by UOW.
Consequently, the complete $\mathrm{FSCL}^{\dagger}$ objective corresponds to a particular UOW objective with $\alpha=0$ and a fixed value of $\lambda$, as formalized in Proposition~\ref{prop:fscl-dagger-equivalence}.

\begin{proposition}[ $\mathrm{FSCL}^{\dagger}$ as a particular UOW instance ]
\label{prop:fscl-dagger-equivalence}
Assume that training uses two normalized views per sample, with the other view as the unique positive associated with each anchor.
Assume further that $\mathrm{FSCL}^{\dagger}$ and UOW use the same conditional samples, leave-one-out denominators, view weights.
If UOW uses $\alpha=0$, the KL--KDE discrepancy associated with $k_\tau$, and a regularization weight satisfying $\lambda/(1-\lambda)=\tau C_{\mathrm{norm}}^{k_\tau}/2$, then there exist constants $a>0$ and $c$, independent of the representations, such that
\begin{equation}
    \mathcal L_{\mathrm{FSCL}^{\dagger}}
    =
    a\,\mathcal L_{\mathrm{UOW}}
    +
    c.
    \label{eq:fscl-dagger-full-equivalence}
\end{equation}
\end{proposition}

\begin{proof}
Expanding the logarithm of the contrastive ratio separates the attractive term from the logarithm of the denominator.
Replacing the denominator sum with an empirical average adds only the logarithm of its cardinality, which is independent of the representation values for a fixed mini-batch.
For two normalized views $x^{(v_1)}$ and $x^{(v_2)}$, the invariance loss satisfies $\mathcal L_{\mathrm{inv}}=\frac{1}{2}\mathbb E[1-(x^{(v_1)})^\top x^{(v_2)}]$.
The attractive component of $\mathrm{FSCL}^{\dagger}$ is therefore equal to $(2/\tau)\mathcal L_{\mathrm{inv}}-1/\tau$.
By~\eqref{eq:fscl-dagger-regularizer}, the repulsive component is equal, up to normalization and an additive constant, to the conditional KL--KDE regularizer used by UOW.
Matching the relative coefficients of the invariance and regularization terms gives $\lambda/(1-\lambda)=\tau\, C_{\mathrm{norm}}^{k_\tau}/2$.
All remaining terms are independent of the representations and can be absorbed into $c$.
\end{proof}

This correspondence applies to the self-supervised $\mathrm{FSCL}^{\dagger}$ objective introduced in~\citep{FSCL}.
The fully supervised FSCL objective additionally uses target labels to construct its positive and negative sets and is therefore better interpreted as a target-restricted form of conditional distributional spreading.

This exact equivalence concerns a specific UOW instantiation that reproduces the conditional KL--KDE estimator used by $\mathrm{FSCL}^{\dagger}$.
Our default spherical UOW implementation instead uses MMD with the Heat kernel $\varphi_t$ and is therefore not identical to $\mathrm{FSCL}^{\dagger}$.
Nevertheless, both objectives follow the same principle of regularizing the representation distribution within each bias group.

\section{Implementation}
\label{app:implementation}

All experiments use AdamW with weight decay $5\times10^{-4}$ and a one-epoch linear warm-up from $1\%$ of the target learning rate, followed by a constant schedule.
Unless stated otherwise, self-supervised methods use two independently augmented views.
Within each dataset, all representation-learning methods share the same encoder, projector, augmentations, batch size, and training budget.

\paragraph{Synthetic Engraving.}
Samples are generated online as $64\times64$ grayscale images from an $8\times8$ binary DataMatrix code and a local noise signature.
The 64 code bits jointly define the observed bias, and each batch of 128 images contains exactly $m\in\{2,4,128\}$ signatures per code.
Augmentations comprise affine transformations, brightness and contrast perturbations, Gaussian blur, uniform noise, random erasing, and normalization to $[-1,1]$.
We use a grayscale-adapted DINOv3 ConvNeXt-Small encoder followed by a 256-dimensional normalized projection head.
Training runs for at most $40$ epochs with learning rate $10^{-4}$.
OW and UOW use KL--KDE with heat-kernel scale parameter $t=2/256$.
Unless stated otherwise, we set $\lambda=1/3$ and $\alpha=0.25$.
Early stopping with patience eight and checkpoint selection maximize validation $\mathrm{mAP}_{|c}$ computed on the encoder representations.

\paragraph{Colored MNIST.}
Images are resized to $32\times32$ and augmented with random affine, brightness, and contrast transformations.
A two-block ConvNet produces 128-dimensional features, followed by a two-layer $512$--$64$ MLP with GELU and Batch Normalization.
We train for $100$ epochs with batch size $256$ and learning rate $0.25$.
The spherical variant uses MMD with a heat kernel of scale parameter $4/64$; the Gaussian variant uses the corresponding Gaussian regularizer.
We set $\lambda=0.05$ and $\alpha=0.5$.
Frozen representations are evaluated using logistic regression and $5$-nearest neighbors.

\paragraph{CelebA and Waterbirds.}
CelebA and Waterbirds use an ImageNet-pretrained ResNet-18 with a 256-dimensional output, followed by a two-layer $1024$--$64$ projector with GELU and Batch Normalization.
Both are trained for $30$ epochs with batch size $256$, learning rate $5\times10^{-2}$, and $\lambda=0.05$.
Spherical variants use MMD with heat-kernel scale parameter $4/64$, whereas Gaussian variants use the induced Gaussian regularizer.
CelebA is loaded from \texttt{flwrlabs/celeba}.
Training images undergo a random resized crop to $128\times128$, horizontal flipping, color jitter, grayscale conversion, and ImageNet normalization; evaluation uses deterministic resizing and center cropping.
We set $\alpha=0.5$ and evaluate \texttt{T=a/S=m} and \texttt{T=b/S=m}.
Frozen logistic regression uses $C=0.01$ and $C=0.001$, respectively.
Waterbirds uses random resized crops to $224\times224$, horizontal flipping, color jitter, grayscale conversion, and ImageNet normalization.
Evaluation uses resizing to $256\times256$ followed by a $224\times224$ center crop.
We use bird type as the target, background as the observed bias, and set $\alpha=0$.

\paragraph{Additional CelebA Protocols.}
The intersectional experiment retains the CelebA architecture and optimization settings, with \emph{Attractive} as the target and \emph{Male} and \emph{Young} as observed biases.
We compare conditioning on \emph{Male}, separate conditioning on both attributes, and joint conditioning on their four intersectional groups.
For from-scratch pretraining, we randomly initialize the ResNet-18 and train it for $200$ epochs with $\lambda=0.05$ and $\alpha=0$ over four seeds.
During training, an online linear probe evaluates \emph{Attractive} accuracy and EO with respect to \emph{Male} on the encoder output $h=f(x)$; its gradients do not update the encoder.
The figure reports means over the four seeds, with epoch 75 marked as a reference checkpoint.

\paragraph{Ablations and Reporting.}
The Synthetic Engraving ablation uses $m=2$, seed 0,
$\lambda\in\{1/4,1/3,1/2\}$, and
$\alpha\in\{0,0.25,0.5,0.75,1\}$.
The CelebA ablation uses
$\lambda\in\{10^{-2},2.5\times10^{-2},5\times10^{-2},
7.5\times10^{-2},10^{-1}\}$ and the same values of $\alpha$.
Its results are averaged over seeds 0, 1, and 2.
Unless stated otherwise, reported means and standard deviations use seeds 0, 1, and 2; Waterbirds additionally uses seed 3.
CelebA and Waterbirds are evaluated with frozen logistic-regression probes, CMNIST with frozen logistic-regression and $5$-NN probes, and Synthetic Engraving with conditional retrieval $\mathrm{mAP}_{|c}$.

\end{document}